\documentclass[12pt]{article}

\newcommand{\hr}{\mathbf{x}^\mathrm{HR}}
\newcommand{\lr}{\mathbf{x}^\mathrm{LR}}
\newcommand{\res}{\mathbf{e_0}}
\newcommand{\xx}{\mathbf{x}}
\newcommand{\bi}{\mathbf{I}}
\newcommand{\bz}{\mathbf{0}}

\usepackage{etoolbox}

\usepackage{url}
\usepackage{setspace}

\usepackage{algorithm}
\usepackage{algpseudocode}
\usepackage{bm}

\usepackage[table,xcdraw]{xcolor}

\usepackage{graphicx}
\usepackage{multirow}
\usepackage{arydshln}

\usepackage{caption}
\usepackage{subcaption}

\usepackage{amsmath} 
\usepackage{amsfonts}
\usepackage{nicefrac}

\usepackage{hyperref}
\hypersetup{
	colorlinks=true,
	linkcolor=blue,
	filecolor=blue,      
	urlcolor=blue,
	citecolor=blue,
}

\usepackage[a4paper, total={6in, 8in}]{geometry}

\usepackage[symbol]{footmisc}

\usepackage{multirow}
\usepackage[numbers,sort&compress]{natbib}

\begin{document}
	\begin{center}	
\sf {\Large {\bfseries MRI super-resolution reconstruction using efficient diffusion probabilistic model with residual shifting}} \\  
		\vspace*{10mm}
		Mojtaba Safari$ ^1 $, Shansong Wang$ ^1 $, Zach Eidex$ ^1 $, Qiang Li$ ^1 $, Richard L.J. Qiu$ ^1 $, Erik H. Middlebrooks$ ^2 $, David S. Yu$ ^1 $, and Xiaofeng Yang$ ^{1, \ddagger} $ \\
	
\end{center}
{$ ^1 $Department of Radiation Oncology and Winship Cancer Institute, Emory University, Atlanta, GA 30322, United States.\\$ ^2 $Department of Radiology, Mayo Clinic, Jacksonville, FL, United States.\\$ ^\ddagger  $Corresponding Author}
\vspace{5mm}\\

\noindent Email: \url{xiaofeng.yang@emory.edu} \\

\newpage

\begin{abstract} 
	\noindent \textbf{Objective:} Magnetic resonance imaging (MRI) is essential in clinical and research contexts, providing exceptional soft-tissue contrast. However, prolonged acquisition times often lead to patient discomfort and motion artifacts. Diffusion-based deep learning super-resolution (SR) techniques reconstruct high-resolution (HR) images from low-resolution (LR) pairs, but they involve extensive sampling steps, limiting real-time application. To overcome these issues, this study introduces a residual error-shifting mechanism markedly reducing sampling steps while maintaining vital anatomical details, thereby accelerating MRI reconstruction. \textbf{Approach:} We developed Res-SRDiff, a novel diffusion-based SR framework incorporating residual error shifting into the forward diffusion process. This integration aligns the degraded HR and LR distributions, enabling efficient HR image reconstruction. We evaluated Res-SRDiff using ultra-high-field brain T1 MP2RAGE maps and T2-weighted prostate images, benchmarking it against Bicubic, Pix2pix, CycleGAN, SPSR, I\textsuperscript{2}SR, and TM-DDPM methods. Quantitative assessments employed peak signal-to-noise ratio (PSNR), structural similarity index (SSIM), gradient magnitude similarity deviation (GMSD), and learned perceptual image patch similarity (LPIPS). Additionally, we qualitatively and quantitatively assessed the proposed framework's individual components through an ablation study and conducted a Likert-based image quality evaluation. \textbf{Main results:} Res-SRDiff significantly surpassed most comparison methods regarding PSNR, SSIM, and GMSD for both datasets, with statistically significant improvements (\textit{p}-values$ \ll 0.05$). The model achieved high-fidelity image reconstruction using only four sampling steps, drastically reducing computation time to under one second per slice. In contrast, traditional methods like TM-DDPM and I\textsuperscript{2}SR required approximately 20 and 38 seconds per slice, respectively. Qualitative analysis showed Res-SRDiff effectively preserved fine anatomical details and lesion morphologies. The Likert study indicated that our method received the highest scores, $4.14 \pm 0.77$(brain) and $4.80 \pm 0.40$(prostate). \textbf{Significance:}Res-SRDiff demonstrates efficiency and accuracy, markedly improving computational speed and image quality. Incorporating residual error shifting into diffusion-based SR facilitates rapid, robust HR image reconstruction, enhancing clinical MRI workflow and advancing medical imaging research. Code available at \url{https://github.com/mosaf/Res-SRDiff}

\end{abstract}

\textbf{\textit{keywords:}} {Super-resolution, MRI, Deep learning, Reconstruction, Diffusion model, Brain T1 map, Ultra-high field MRI}

\newpage 
\section{Introduction}

Magnetic resonance imaging (MRI) is an indispensable tool in both clinical practice and research, providing detailed anatomical and functional images. Quantitative techniques, such as 3D magnetization-prepared 2 rapid acquisition gradient echo (MP2RAGE) T1-maps, offer robust imaging free from reception bias and first-order transmit field inhomogeneities, thereby enabling precise diagnosis and treatment planning~\cite{MARQUES20101271, doi:10.1148/radiol.2016141802, doi:10.1016/j.jcmg.2015.11.005}. For example, T1-maps are employed to identify hypoxic regions that can inform adaptive dose-painting radiation therapy~\cite{doi:10.1148/radiol.12120777, EPEL2019977, https://doi.org/10.1002/mp.17353}. Moreover, in addition to these quantitative methods, T2-weighted (T2w) MRI provides enhanced tissue contrast, rendering it a critical imaging modality for prostate cancer treatment by delineating tumor boundaries and guiding therapeutic decisions~\cite{https://doi.org/10.1002/mp.14196}. Nevertheless, the lengthy acquisition times associated with both T1-mapping and T2w imaging may induce patient discomfort and elevate the risk of motion artifacts~\cite{Safari_2024_MAUDGAN}, thereby potentially compromising image quality and diagnostic accuracy.

To accelerate MRI image acquisition, super-resolution (SR) studies have aimed to reconstruct high-resolution (HR) images from their low-resolution (LR) counterparts~\cite{LEPCHA2023230}. Conventional SR models, which constitute a subcategory of the broader field of image restoration, employ a maximum a posteriori framework--a Bayesian paradigm consisting of a likelihood (loss) function and a prior (regularization) term--to resolve the ill-posed SR task. The likelihood term presupposes an underlying noise distribution, yielding $\ell_2$ and $\ell_1$ losses for Gaussian and Laplacian noise assumptions, respectively. Typical regularizers include Tikhonov~\cite{10.1007/978-3-540-79490-5_8}, non-local similarity~\cite{6392274}, wavelet~\cite{4770145}, and total variation~\cite{5193008} to address the ill-posed image restoration task.

Deep learning algorithms, particularly generative deep learning models, consistently outperform traditional methods in medical imaging tasks such as reconstruction~\cite{https://doi.org/10.1002/mp.17675, 9703109} and denoising~\cite{9340274}. Despite the impressive visual fidelity achieved by generative adversarial networks (GANs), they often face challenges such as mode collapse and unstable training~\cite{YI2019101552, 10738507}, which might undermine their reliability in practical and clinical settings. For example, Cheng \textit{et al.} introduced a structure-preserving SR (SPSR) method that incorporates gradient guidance into the SR process~\cite{Ma_2020_CVPR}. This approach highlights the importance of high-fidelity gradient maps in preserving geometric consistency and mitigating structural distortions, which are prevalent challenges in GAN-based super-resolution techniques.

Recently, diffusion models have emerged as a compelling alternative to address these limitations. These models have demonstrated considerable success in MRI-related applications, such as reconstruction~\cite{10.1117/12.3002863}, denoising~\cite{Pan_2023}, synthesis~\cite{10.1117/12.3047506}, and super-resolution~\cite{Chang_2024, 10737883}. The operational framework of diffusion models involves a forward process that gradually diffuses data towards a prior distribution, typically modeled as a multivariate standard Gaussian ($ \mathcal{N}(\bz, \bi) $), followed by a reverse process in which a neural network (NN) is trained to approximate the inverse trajectory across numerous sampling steps. However, two significant drawbacks associated with the standard paradigm for image SR persist: first, the iterative nature of the denoising process renders the generation of HR images from LR inputs computationally demanding; second, the reliance on pure Gaussian noise for initializing the reverse process is inherently more suited for image generation than for restoration tasks. Yue \textit{et al.}~\cite{yue2023resshift} highlighted these inefficiencies, and subsequent research has demonstrated that centering the initial reconstruction distribution around the LR image--by adjusting residual errors over $ T $ sampling steps--can significantly enhance sampling efficiency~\cite{10681246}.

Recent advancements in diffusion models have significantly reduced the required sampling steps, enhancing their efficiency. For example, a simulation-free Image-to-Image Schr\"{o}dinger Bridge (I\textsuperscript{2}SB) framework~\cite{liu2023schrodinger} employs a nonlinear diffusion bridge that directly utilizes degraded image information to guide restoration, resulting in more interpretable generative pathways. In parallel, partial diffusion models~\cite{10720924} have been introduced for MRI applications, leveraging latent convergence observations between LR and HR images. This latent alignment strategy effectively bypasses redundant denoising steps, markedly decreasing computational load. Additionally, a method that distills the stochastic diffusion process into a single deterministic generation step~\cite{Wang_2024_CVPR} has been proposed, achieving substantial acceleration without sacrificing perceptual quality.

In this study, we present an efficient diffusion model named \textbf{Res-SRDiff}, which leverages the residual error shift between HR and LR image pairs to reconstruct HR axial T2w prostate images and quantitative brain MRI T1 MP2RAGE maps obtained from ultra-high B\textsubscript{0} fields, extending the work presented in~\cite{yue2023resshift,10681246}. To our knowledge, this is the first investigation aimed at recovering HR MRI using an efficient diffusion model that requires \textbf{only four sampling steps}, in contrast to the thousands required by conventional diffusion models. This considerable reduction in sampling steps substantially enhances computational efficiency while maintaining the high quality of the restored HR images.

The contributions of this work are:

\begin{itemize} 
	\item Formulating an efficient diffusion model specifically tailored for SR task, enabling inference in only four sampling steps. 
	\item Utilizing a U-net architecture integrated with a Swin Transformer block, replacing the traditional attention layer, to ensure improved generalization across varying image resolutions.
	\item Conducting extensive evaluations using publicly available axial T2w prostate image datasets and institutionally acquired ultra-high-field (7T) T1 MP2RAGE brain MRI maps.
	\item Demonstrating, for the first time, the application of an efficient diffusion model to reconstruct HR axial T2w pelvic images and ultra-high B\textsubscript{0} field brain T1 maps from LR image pairs.
\end{itemize}

\section{Materials and Methods}

In this section, we first review the traditional Denoising Diffusion Probabilistic Model (DDPM). Next, we introduce our proposed method, Res-SRDiff, which is designed to recover HR images ($  {\hr} $) from their LR counterparts ($  {\lr} $). We assume that both HR and LR images have similar spatial size, an assumption that can be readily satisfied by pre-upsampling the LR images using nearest neighbor interpolation.

\subsection{DDPM} 

The DDPM was initially inspired by non-equilibrium thermodynamics~\cite{pmlr-v37-sohl-dickstein15}, aiming to approximate a complex data distribution with a tractable distribution, such as a standard Gaussian distribution. It was later enhanced by integrating stochastic differential equations and denoising score matching~\cite{song2021scorebased,10.5555/3495724.3496298}. The DDPM comprises two diffusion processes: a forward process and a reverse process. The forward process degrades the input image into noise following a standard Gaussian distribution $ \mathcal{N}(\bz, \bi) $ over numerous steps $T$. The reverse process trains an NN to approximate the sampling trajectory required to recover the input image from Gaussian noise over a large number of steps $T$, which diminishes the sampling efficiency of the DDPM.

\subsection{Problem formulation}

\begin{figure}[t] 
	\centering 
	\includegraphics[width=\textwidth]{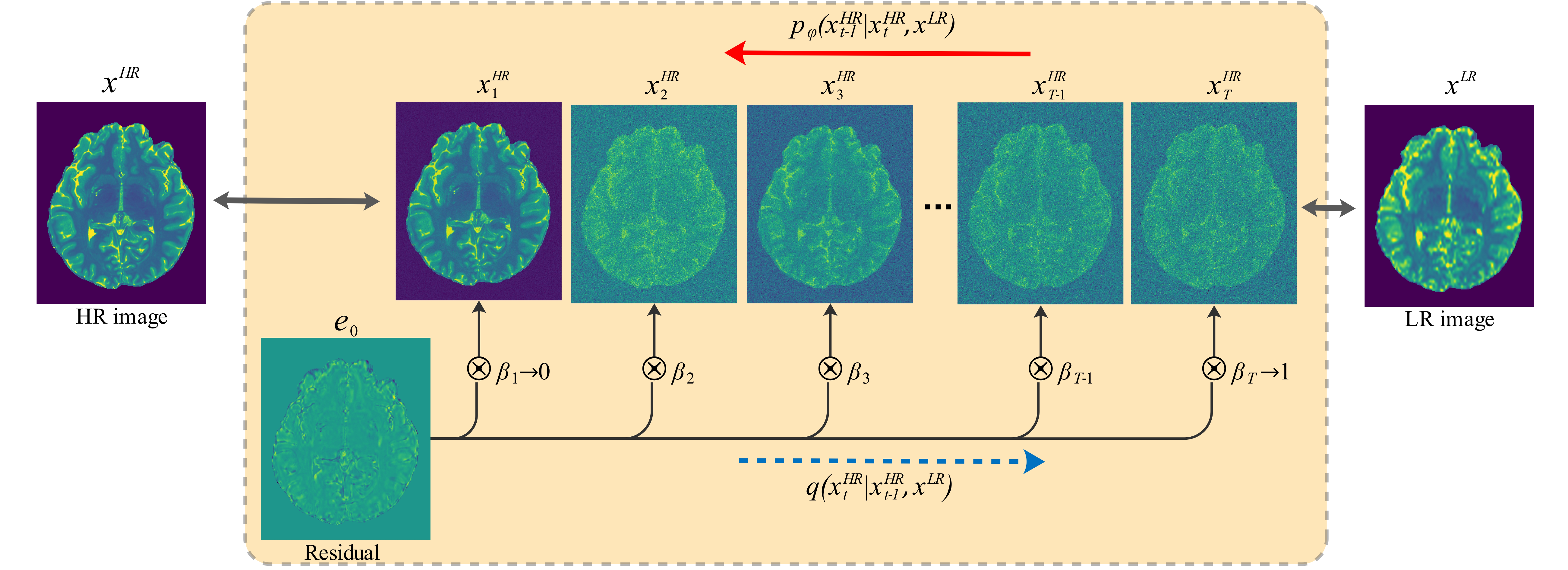} 
	\caption{Illustration of the diffusion process. The forward diffusion process in Res-SRDiff, where a HR image $ {\hr}$ is progressively shifted to match the LR distribution $q( {\lr})$. The model introduces a residual error $e_0 =  {\lr} -  {\hr}$, which drives $ {\hr}$ through $T$ Markov steps until $q({\hr_T}) \approx \mathcal{N}(\lr, \gamma \bi)$, rather than converging to a standard Gaussian distribution. The reverse process trains an NN $g_\varphi$ to estimate the posterior distribution $ p_\varphi (\hr \vert \lr)$ given in \eqref{eq:posterior_01}.} 
	\label{fig:diffusoin_processes} 
\end{figure}

Res-SRDiff is built upon a Markov chain, similar to the conventional DDPM model. However, it aims to degrade input HR images $ {\hr}$ into an image $\hr_T$ over $T$ steps such that the resulting distribution $q({\hr_T}) \approx \mathcal{N}(\lr, \gamma \bi)$ rather than converging to $\mathcal{N}(\bz, \bi)$. This is achieved by introducing the residual $\res =  {\lr} -  {\hr}$, which is used to shift $ {\hr}$ over the $T$ steps. This process is illustrated in \figurename~\ref{fig:diffusoin_processes}.

\paragraph{Forward process.} To simulate the forward diffusion process, a monotonically increasing shifting sequence $ \{\beta_t\}^T_{t=1}$ over time steps $t$ with bounding conditions $\beta_1 \to 0$ and $\beta_T \to 1$ is used. The transition kernel for simulating the forward diffusion process is given in \eqref{eq:forward_process}, which is constructed based on the Markov chain and the residual error $\res$ shift sequences (see \figurename~\ref{fig:diffusoin_processes}):

\begin{equation}
	\centering
	q(\hr_t \vert \hr_{t-1},  {\lr}  ) = \mathcal{N}( \hr_t; {\hr_{t-1}} + \res \alpha_t, \gamma^2 \alpha_t \bi ), \,\,\,\,\ t \in [1, T]
	\label{eq:forward_process}
\end{equation}
where $\alpha_1 = \beta_1 \to 0$ and $\alpha_t = \beta_t - \beta_{t-1}$ for $t > 1$, and $\gamma$ is a hyper-parameter introduced to improve the flexibility of the forward diffusion process. Considering the Markov chain, we can compute the image at step $t$ from its step $t-1$ using the reparameterization trick as follows:

\begin{equation}
	\centering
	{\hr_t} = \hr_{t-1} + \alpha_t \res + \sqrt{\gamma^2 \alpha_t} \bm{\epsilon}
	\label{eq:forward_process_v1}
\end{equation}
where $\bm{\epsilon} \sim \mathcal{N}(\bz, \bi)$. Since this sampling forward process using (\ref{eq:forward_process_v1}) increases the computational burden, it is also possible to compute the image at step $t$ directly from the input noise-free image as follows:

\begin{subequations} 
	\begin{align}
		\xx_1 &= \xx_0 + \alpha_1 \res + \sqrt{\gamma^2 \alpha_1} \bm{\epsilon} \label{eq:forward_diffusion_a} \\
		\xx_2 &= \xx_1 + \alpha_2 \res   + \sqrt{\gamma^2 \alpha_2} \bm{\epsilon} \label{eq:forward_diffusion_b} \\
		&= \xx_0 + (\alpha_1 + \alpha_2) \res + \gamma \left( \sqrt{\alpha_1} + \sqrt{\alpha_2} \right) \bm{\epsilon} \label{eq:forward_diffusion_c} \\
		&\vdots \nonumber \\ 
		\xx_t &= \xx_0 + \res \sum_{t^\prime = 1}^{T} \alpha_{t^\prime}  + \gamma (\sum_{t^\prime = 1}^{T} \sqrt{\alpha_{t^\prime}}) \bm{\epsilon} \label{eq:forward_diffusion_d}
	\end{align}
\end{subequations} 

Here, we omit the superscript HR for brevity and $\xx_0$ is $ {\hr}$. The second term (mean) and the square of the third term (variance) in the summation given in \eqref{eq:forward_diffusion_d} are equal to $\beta_t$. Thus, the marginal distribution at any time step $t$ can be computed analytically as follows:

\begin{equation}
	\centering
	q(\hr_t \vert  \hr,  \lr, t ) = \mathcal{N}( \hr_{t};  \hr + \res \beta_t, \gamma^2 \beta_t \bi ), \,\,\,\,\ t \in [1, T]
	\label{eq:complete_forward_process}
\end{equation}

\paragraph{Reverse process.} The reverse process trains a NN $g_\varphi$ to estimate the posterior distribution $p_\varphi( {\hr} \vert  {\lr})$, as follows~\cite{luo2022understandingdiffusionmodelsunified}:

\begin{equation}
	p_\varphi(\hr \vert \lr) = \int  p(\hr_T \vert \lr) \prod_{t=1}^{T} p_\varphi (\hr_{t-1} \vert \hr_t, \lr) d\xx_{1:T}
	\label{eq:posterior_01}
\end{equation}
where $p(\hr_T \vert \lr) \approx \mathcal{N} (\lr, \gamma^2 \bi)$ and $p_\varphi (\hr_{t-1} \vert \hr_t, \lr)$ is a reverse transition kernel that aims to learn $\hr_{t-1}$ from $\hr_t$ by training a network $g_\varphi$. Similar to conventional diffusion models~\cite{10.5555/3495724.3496298, song2021scorebased, luo2022understandingdiffusionmodelsunified}, it can be written as follows by adopting the Gaussian assumption:
\begin{equation}
	p_\varphi (\hr_{t-1} \vert \hr_t,  \lr) = \mathcal{N}(\hr_{t-1}; \bm{\mu_\varphi}(\hr_t, \lr, t), \bm{\Sigma_\varphi}(t))
	\label{eq:gasussian_assumption_02}
\end{equation}
where the optimum parameter $\varphi$ is achieved by minimizing the Kullback-Leibler (KL) divergence between the forward and reverse kernels summed over all time steps as follows~\cite{luo2022understandingdiffusionmodelsunified}:

\begin{equation}
	 \underset{\varphi}{\arg \min} \sum_{t} D_{KL} \left(q({\hr_{t-1}} \vert {\hr_{t}},  \hr,  {\lr}  ) \vert p_\varphi ( \hr_{t-1} \vert  \hr_t,  \lr)\right)
	 \label{eq:kl_divergence}
\end{equation}

The target distribution $q({\hr_{t-1}} \vert {\hr_{t}},  \hr,  {\lr})$ can be computed using \eqref{eq:forward_process} and \eqref{eq:complete_forward_process}, along with the Markov chain assumption, which states $\hr_t \perp \hr_{1:t-2} \vert \hr_{t-1}$, as follows:

\begin{equation}
	\centering
	\begin{aligned}
		q(\hr_{t-1} \vert \hr_t,  \hr,  \lr  ) &= q(\hr_{t} \vert \hr_{t-1}, \lr ) q(\hr_{t-1} \vert  \hr, \lr)\\
		&=\mathcal{N}(\hr_{t} ; \hr_{t-1} + \alpha_t \res,\gamma^2 \alpha_t \bi) \mathcal{N}( \hr_{t-1};  {\hr} + \res \beta_{t-1}, \gamma^2 \beta_{t-1} \bi )
	\end{aligned}
\end{equation}

The multiplication of two Gaussian distributions yields another Gaussian distribution that can be computed tractably~\cite{pml1Book} as follows:

\begin{equation} 
	\centering
	q( \hr_{t-1}\vert \hr_{t},  \hr,  {\lr}) = \mathcal{N} (    \hr_{t-1} ; \underset{\bm{\mu_q}}{\underbrace{\dfrac{\beta_{t-1}}{\beta_{t}}  \hr_t + \dfrac{\alpha_t}{\beta_t}  \hr}}, \underset{\bm{\Sigma_q}}{\underbrace{\gamma^2\alpha_t \dfrac{\beta_{t-1}}{\beta_{t}}\mathbf{I}}}  )
	\label{eq:margianl_prob_02}
\end{equation}

By assuming that the forward and backward covariance matrices are similar ($\bm{\Sigma_q} = \bm{\Sigma_\varphi}$), the KL divergence given in \eqref{eq:kl_divergence} simplifies to:
\begin{equation}
	\centering
	\begin{aligned}
		\hat{\varphi} &= \underset{\varphi}{\arg \min} \frac{1}{2} \left[ (\bm{\mu_\varphi} - \bm{\mu_q})^T\bm{\Sigma_q(t)}^{-1}(\bm{\mu_\varphi} - \bm{\mu_q})\right]\\
				&=\underset{\varphi}{\arg \min} \frac{1}{2}\left[ (\bm{\mu_\varphi} - \bm{\mu_q})^T\dfrac{\beta_{t} \mathbf{I}}{\gamma^2\alpha_t \beta_{t-1}}(\bm{\mu_\varphi} - \bm{\mu_q})\right]\\
				&=\underset{\varphi}{\arg \min} \frac{\beta_{t} }{2 \gamma^2\alpha_t \beta_{t-1}}\left[ \parallel \bm{\mu_\varphi} - \bm{\mu_q} \parallel^2_2\right]
	\end{aligned}
	\label{eq:kl_divergence_02}
\end{equation}
The mean parameter $\bm{\mu_\varphi}( \hr_t,  \hr,  \lr, t) $ is parameterized as follows:

\begin{equation}\label{eq:sampling_v01}
	\centering
	\bm{\mu_\varphi}( \hr_t,  \hr,  \lr, t) =\dfrac{\beta_{t-1}}{\beta_{t}}  \hr_t + \dfrac{\alpha_t}{\beta_t} g_\varphi(\hr_t,  \lr ,t)
\end{equation}

After substituting it into \eqref{eq:kl_divergence_02}, the final loss function is achieved as follows:

\begin{equation}
	\hat{\varphi} = \underset{\varphi}{\arg \min} \parallel g_\varphi(\hr_t,  \lr ,t) -  \hr\parallel_2^2
	\label{eq:mse_loss}
\end{equation}

The constant parameters were dropped, as experiments demonstrated that this improves the model's performance~\cite{10681246, 10.5555/3495724.3496298}. In addition to the data fidelity $\ell_2$ loss, a learned perceptual image patch similarity (LPIPS) $\ell_p$ loss~\cite{Zhang_2018_CVPR} was employed. The overall optimization function is given by:

\begin{equation}\label{eq:total_loss}
	\mathcal{L}_\varphi = \lambda \parallel g_\varphi(\hr_t,  \lr ,t) -  \hr\parallel_2^2 +  \ell_p(g_\varphi(\hr_t,  \lr ,t),  \hr)
\end{equation}
where $\lambda$ is a hyper-parameter controlling the relative importance and we set it to 10 in this study. The training and sampling pseudo-codes are provided in Algorithm~\ref{alg:training_process} and \ref{alg:sampling_process}.

\begin{algorithm}[tbhp]
	\caption{Training process}\label{alg:training_process}
	\begin{algorithmic}
		\Require High-resolution dataset $\mathcal{T}_{HR}$, Low-resolution dataset $\mathcal{T}_{LR}$
		\Repeat 
		
		$ \hr \sim \mathcal{T}_{HR}$, $ \lr \sim \mathcal{T}_{LR} $
		
		$t \sim $ Uniform$(\{1, ..., T\})$
		
		$ \hr_t \sim q( \hr_t \vert  \hr,  \lr,t)$\Comment{Given in \eqref{eq:complete_forward_process}}
		
		Take a gradient descent step on $\nabla\mathcal{L}_\varphi( \hr_t,\lr,t)$\Comment{Given in \eqref{eq:total_loss}}
		\Until
	\end{algorithmic}
\end{algorithm}

\begin{algorithm}[tbhp]
	\caption{Sampling process}\label{alg:sampling_process}
	\begin{algorithmic}
		\Require Low-resolution image $ \lr$
		
		$ \lr_T \sim \mathcal{N}( \lr_T;  \lr, \gamma^2 \beta_T \bi)$
		
		\For{t = T, ..., 1}
		
		$\bm{\epsilon} \sim \mathcal{N}(\bm{\epsilon}; \bz, \bi)$ if $t > 1$ else $\bm{\epsilon} = 0$
		
		$ \bm{\mu_\varphi} =  \dfrac{\beta_{t-1}}{\beta_t}  \hr_t + \dfrac{\alpha_t}{\beta_t} g_\varphi(\hr_t,  \lr ,t)  $ \Comment{Given in \eqref{eq:sampling_v01}}
		
		$ \hr_{t-1} = \bm{\mu_\varphi} + \gamma \sqrt{\dfrac{\beta_{t-1} \alpha_t}{\beta_t}} \bm{\epsilon} $
		\EndFor
		
	\end{algorithmic}
\end{algorithm}

\subsection{Noise scheduler}

\begin{figure}
	\centering
	\includegraphics[width=.9\textwidth]{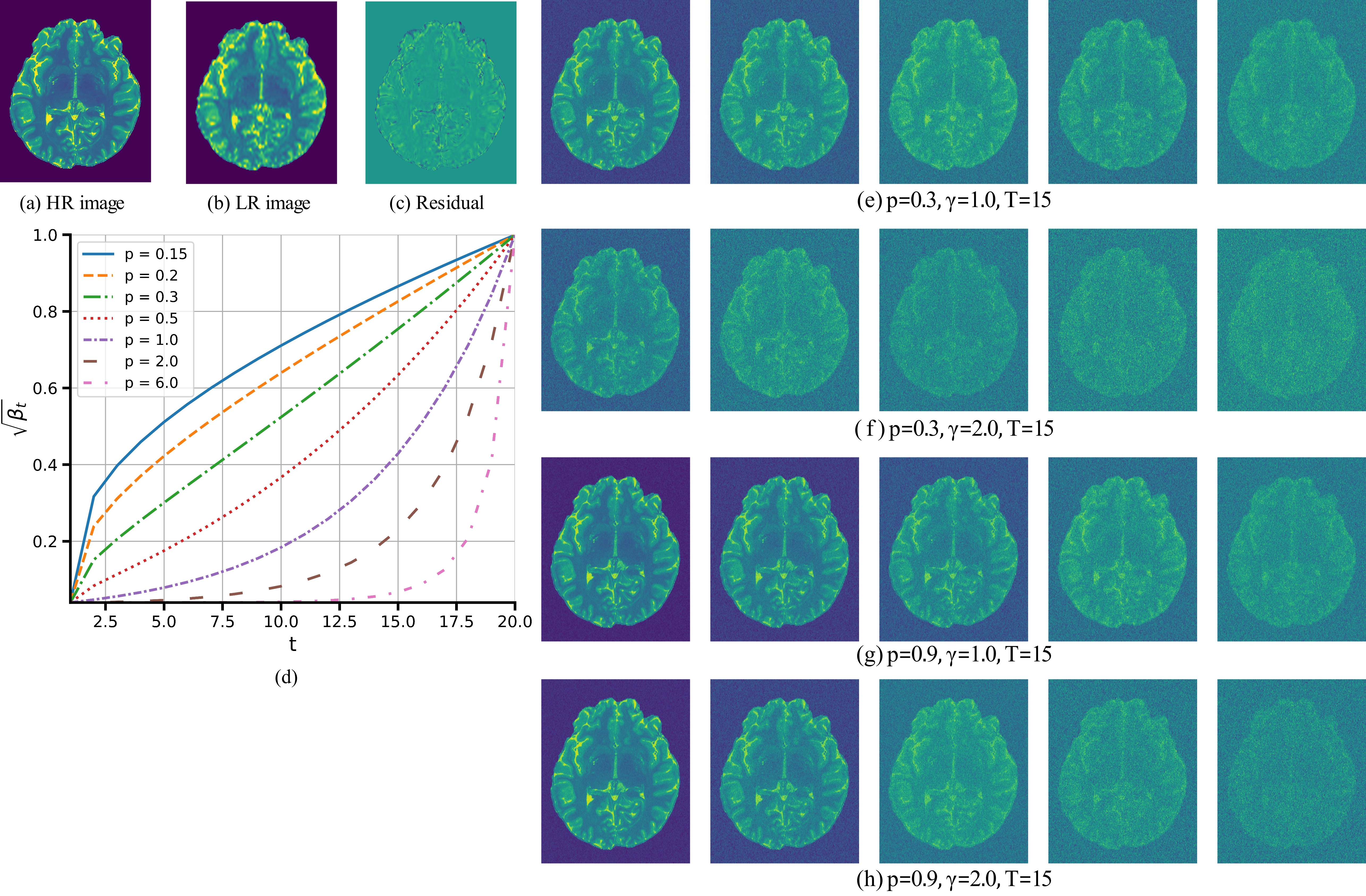}
	\caption{Residual shift denoising diffusion process. (a) shows the HR image, $ \hr $; (b) displays the corresponding LR image, $ \lr $; and (c) illustrates the residual error, $ \res =  \lr -  \hr $. (d) presents the evaluation of the noise scaling factor, $ \sqrt{\beta_t} $, as a function of the diffusion time step, $ t $. Panels (e)-(h) demonstrate the forward diffusion process driven by the residual error shift for different hyper-parameter sets.}
	\label{fig:hyper_parameters_p}
\end{figure}

This study utilizes a hyper-parameter $\gamma$ and a noise scheduler $\{\beta_t\}_{t=1}^T$ in the forward diffusion process. Given that $\sqrt{\beta_t}$ and the scaling factor $\gamma$ in \eqref{eq:complete_forward_process} control the forward process, and it has been shown that a NN can approximate the forward diffusion trajectory~\cite{pmlr-v37-sohl-dickstein15, 10.5555/3495724.3496298}, $\gamma \sqrt{\beta_1}$ needs to be small such as 0.04, which ensures that $q(\hr_1\vert  \hr,  \lr) \approx q( \hr)$. Thus, we set $\beta_1 = (0.04 / \gamma)^2$ and used $\gamma = 2$ to satisfy the first bounding condition $\beta_1 \to 0$ (see \figurename~\ref{fig:diffusoin_processes}) and $\beta_T = 0.9999$ to satisfy the second bounding condition $\beta_T \to 1$. We employed a non-uniform geometric noise scheduler proposed by \cite{10681246} for $\sqrt{\beta_t}$ as follows:

\begin{equation}
	\sqrt{\beta_{t}} = \sqrt{\beta_1} \exp\left[(\dfrac{t-1}{T-1}) ^ p\log\sqrt{\dfrac{\beta_T}{\beta_1}}\right],\,\,\, t \in [2, T-1]
\end{equation}
where the hyper-parameter $p$ controls the growth rate, as shown in \figurename~\ref{fig:hyper_parameters_p}. We used $p=0.3$ in our study, similar to a recent study~\cite{10681246}. Furthermore, we used 15 steps for training and four steps for sampling. 


Figure~\ref{fig:hyper_parameters_p} comprises several panels demonstrating the effects of the hyper-parameters $ p $ and $ \gamma $ on the forward diffusion process. Specifically, panels (a)-(c) show the HR image, the LR image, and the residual error, respectively. Panel (d) illustrates how $ \sqrt{\beta_t} $ varies with the time step $ t $ for different values of $ p $. Finally, panels (e)-(h) depict representative outputs of the forward diffusion process at selected time steps $ t $, enabling a visual comparison of noise levels under varying $ p $ and $ \gamma $.

From panels (e) and (g) as well as (f) and (h), one observes that, for a fixed time step $ t $ and fixed $ \gamma $, reducing $ p $ leads to an increase in the amount of noise in the reconstructed images. This outcome aligns with panel (d), where lower $ p $ corresponds to higher $ \sqrt{\beta_t} $, implying more additive noise given in \eqref{eq:complete_forward_process}. Conversely, keeping $ t $ and $ p $ fixed but increasing $ \gamma $ also produces noisier reconstructions, as evidenced by comparing panels (e) and (f), and similarly (g) and (h).

The Res-SRDiff model was implemented using \texttt{PyTorch} (version 2.5.1) and executed on NVIDIA A100 GPUs. The model was trained for 182,000 and 131,000 steps on the brain and prostate datasets, respectively, using a batch size of 16. The network was optimized with the Rectified Adam (RAdam) optimizer~\cite{liu2021varianceadaptivelearningrate} and employed a cosine annealing learning rate scheduler~\cite{loshchilov2017sgdrstochasticgradientdescent}. The initial learning rate was set in the range of $2 \times 10^{-5}$ to $5 \times 10^{-5}$, and it was adjusted according to a cosine decay schedule throughout training. A warm-up phase of 5,000 steps was applied before transitioning to the cosine decay schedule to stabilize early training dynamics.

\subsection{Patient data acquisition and data preprocessing}

We used institutional ultra-high 7T brain T1 MP2RAGE maps~\cite{Middlebrooks_2024_data} and publicly available axial T2w prostate cancer data~\cite{10.1117/1.JMI.5.4.044501} to train and evaluate the proposed method.

Our institutional dataset comprises 142 patients with a confirmed diagnosis of multiple sclerosis, which were divided into two non-overlapping sets: a training set (121 cases, 14,566 slices) and a test set (21 cases, 2,552 slices). This retrospective study was approved by the Mayo Clinic IRB. The institutional data were acquired using a 7 T Siemens MAGNETOM Terra with 8-channel transmit/32-channel receive head coil with the following key imaging parameters: TR = 4.5 s, TE = 2.2 ms, TI1/TI2 = 0.95/2.5 s, FA1/FA2 = 6$^\circ$/4$^\circ$, FOV = $230\times230$ cm$^2$, matrix size of $288\times288$, a resolution of $0.8 \times 0.8 \times 0.8$ mm$^3$, and a total acquisition time of 8:44 min. FSL BET~\cite{https://doi.org/10.1002/hbm.10062} was used to extract the brain mask from image inversion 1, which was subsequently applied to the T1 maps to remove the noisy background and skull. The T1 maps were down-sampled by a factor of $4^3$, resulting in a voxel size of $3.2 \times 3.2 \times 3.2$ mm$^3$ (a 4-fold reduction in each direction).

We randomly selected data from 334 patients in the public prostate dataset, which were split into two non-overlapping sets: a training set (268 patients, 10,480 slices) and an evaluation set (66 patients, 2,668 slices). The T2w MR images were acquired using a 1.5 T Siemens scanner with the following parameters: TR = 2.2 s, TE = 202 ms, FA = 110$^\circ$, matrix size of $256 \times 256$, an in-plane resolution of $0.66 \times 0.66$ mm$^2$, and a slice thickness of 1.5 mm. The T2w MR images were down-sampled by a factor of 18, yielding a voxel size of $2 \times 2 \times 3$ mm$^3$ (a 9-fold reduction in-plane and a 2-fold reduction along the slice axis).

Under-sampling of  ultra-high B\textsubscript{0} brain T1 maps and the axial T2w prostate images were performed in image space using the \texttt{SimpleITK.Resample} (version 2.1.1) package~\cite{Yaniv2018}.

\subsection{Quantitative and statistical analysis}\label{sec:quant_stat_v01}

We evaluated our method against four benchmark approaches: Bicubic, Pix2pix~\cite{isola2017image}, CycleGAN~\cite{CycleGAN2017}, SPSR~\cite{Ma_2020_CVPR}, I\textsuperscript{2}SR~\cite{liu2023schrodinger}, and TM-DDPM, which is a conventional DDPM with a vision transformer backbone~\cite{Pan_2023}. All methods were trained for the same number of steps and with similar training parameters, except that the TM-DDPM and I\textsuperscript{2}SB model had approximately three times as many training parameters.

The reconstructed HR image quality was quantitatively evaluated using four metrics: peak signal-to-noise ratio (PSNR), structural similarity index (SSIM)\cite{1284395}, gradient magnitude similarity deviation (GMSD)\cite{6678238}, and LPIPS~\cite{1576816}. Higher SSIM and PSNR values, and lower GMSD and LPIPS values, indicate better image restoration performance. PSNR quantifies the residual error between the restored and ground truth images, and its logarithmic scale aligns better with human perceptual judgments~\cite{Safari2023_medfusiongan}. Furthermore, SSIM, GMSD, and LPIPS provide measures of the structural similarity between the restored images and the HR ground truth images.

Two statistical tests were employed to assess the significance of differences: a one-way analysis of variance (ANOVA) and Tukey's honestly significant difference (HSD) test. Prior to these analyses, the Shapiro-Wilk test was conducted to evaluate the normality of the residuals. When the normality assumption was not satisfied, non-parametric methods were used, specifically the Kruskal-Wallis test followed by Dunn's test with Bonferroni correction for multiple comparisons. The ANOVA tested the null hypothesis that the mean values for each method are equal, while the Kruskal-Wallis test assessed whether the distributions of the groups differed significantly. Tukey's HSD and Dunn's test with Bonferroni correction were then used to identify which specific pairs of groups differed significantly. For all analyses, the significance level was set at $p < 0.05$.

\subsection{Subjective Image-Quality Evaluation (Likert Ratings)}
A subjective image-quality evaluation was performed by two experienced medical physicists certified by the American Board of Radiology. Each rater independently reviewed SR images and assigned scores on a 5-point Likert scale ($1  =  $Poor, $ 2 =  $Fair, $ 3 =  $Acceptable, $ 4 =  $Good, $ 5 =  $Excellent) with respect to three criteria: contrast, edge sharpness, and preservation of anatomical detail.  

For every reconstruction method, 25 images were randomly selected from the brain dataset and 25 from the prostate dataset, yielding 50 images per method and 300 images in total. Mean $ \pm $ standard‑deviation scores are reported; larger values denote superior perceived quality. The 95\% confidence intervals (CIs) for the mean values were computed using the percentile bootstrap method with 10,000 iterations, along with the bias-adjusted accelerated bootstrap approach~\cite{2020SciPy-NMeth}. Statistical comparisons between methods were performed using the framework detailed in Section~\ref{sec:quant_stat_v01}, and the resulting \textit{p}-values are provided alongside the descriptive statistics.

\section{Results}

\subsection{Brain T1 maps}

\begin{figure}[t]
	\centering
	\includegraphics[width=\textwidth]{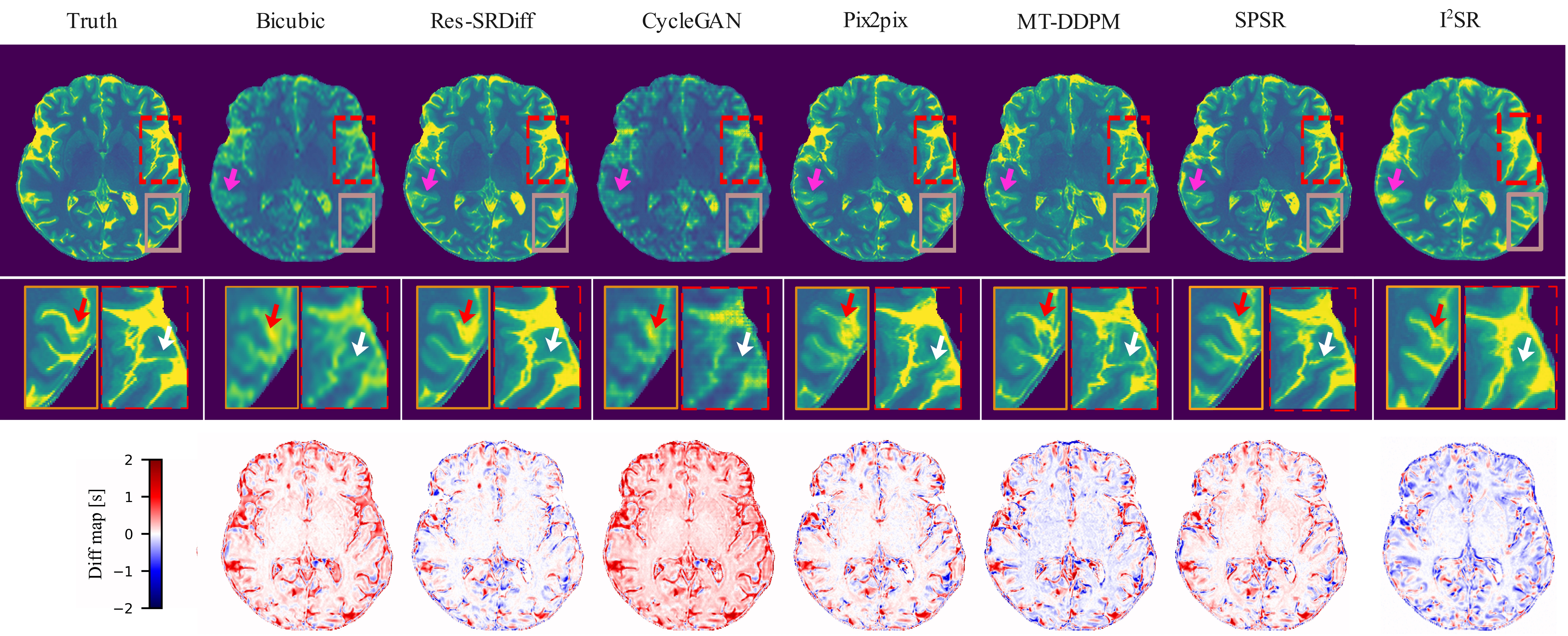}
	\caption{Qualitative results of the ultra-high field brain T1 MP2RAGE maps. The first row shows the ground truth image along with the restored outputs from our proposed Res-SRDiff and comparative models. The second row displays the zoomed-in regions corresponding to the dashed red and brown boxes. The white and red arrows highlight regions where our method outperforms the comparative models. The last row presents the difference map between the restored images and the ground truth.}
	\label{fig:brain_figure_qualitative}
\end{figure}

The proposed method demonstrated superior performance, exhibiting lower residual errors and higher structural similarity, as illustrated in \figurename~\ref{fig:brain_figure_qualitative} (first row). Detailed comparisons provided by the zoomed-in panels (second row of \figurename~\ref{fig:brain_figure_qualitative}), indicated by white and red arrows, reveal that our method effectively captured fine structural details more accurately than baseline methods. However, despite successfully reconstructing these highlighted fine details, our method, like the comparative approaches, was unable to fully reconstruct very subtle structures diminished by partial volume effects, as indicated by the pink arrow in \figurename~\ref{fig:brain_figure_qualitative}. The difference maps shown in the third row of \figurename~\ref{fig:brain_figure_qualitative} further confirm the lower global discrepancies between our method's outputs and the HR ground truth images. These visual observations align well with quantitative findings, where our proposed method yielded higher PSNR values and lower GMSD and LPIPS scores (refer to \tablename~\ref{tab:qmetrics}). The reduced global residual errors depicted in the third row of \figurename~\ref{fig:brain_figure_qualitative} additionally suggest a greater consistency between the reconstructed outputs of our method and the ground truth images.

In terms of computational efficiency, the proposed method achieved an average evaluation time of $0.46 \pm 0.21$ seconds per slice, which was substantially faster than I\textsuperscript{2}SR ($31.15 \pm 0.27$ seconds per slice) and the MT-DDPM method ($66.84 \pm 27.72$ seconds per slice). Given that the quantitative metrics did not meet the Shapiro-Wilk normality test criteria (\textit{p}-values $\ll 0.001$), non-parametric analyses, specifically Kruskal-Wallis and Dunn's post-hoc tests, were conducted. The Kruskal-Wallis test indicated significant statistical differences among the evaluated methods (\textit{p}-values $\ll 0.0001$) across all quantitative metrics. On average, the proposed method consistently outperformed alternative approaches across PSNR, GMSD, and LPIPS with statistically significant improvements (\textit{p}-values $\ll 0.001$), except in the case of LPIPS, where the difference compared to the Pix2pix method was not statistically significant (\textit{p} $= 0.08$).

The Shapiro-Wilk test indicated that the Likert scores were not normally distributed (\textit{p} $ \ll 0.001 $) for all evaluated methods. Our proposed method achieved an average score of $4.14 \pm 0.77$ (95\% CI 3.90, 4.31), outperforming the second-best method, TM-DDPM, which had an average score of $3.51 \pm 0.75$ (95\% CI 3.29, 3.71); however, this difference was not statistically significant (\textit{p} $ =0.11 $). In contrast, the differences between these two top-performing methods and the third-best method, Pix2pix with score $3.04 \pm 0.68$ (95\% CI 2.84, 3.20), were statistically significant (\textit{p} $ \ll 0.001 $).

\subsection{Pelvic T2w images}

\begin{figure}[t]
	\centering
	\includegraphics[width=\textwidth]{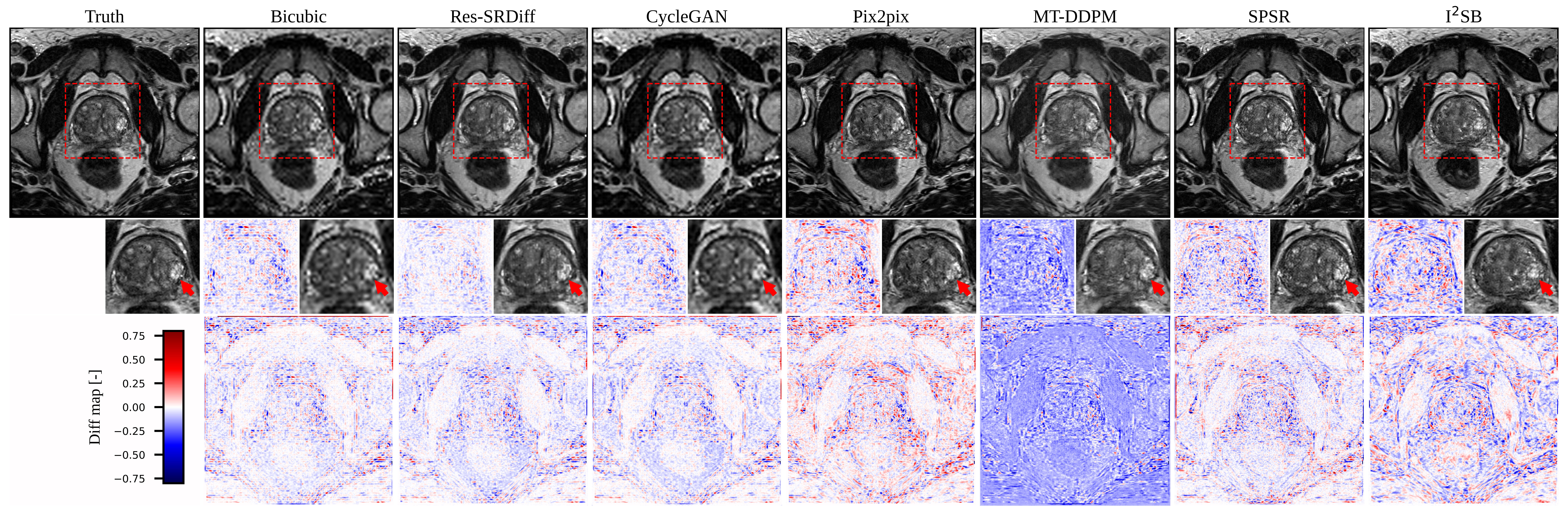}
	\caption{Qualitative results of the pelvic axial T2w images. The first row presents the ground truth image along with the restored outputs from our proposed Res-SRDiff and comparative models. The second row shows the zoomed-in regions outlined by the red dashed lines, where the red arrows indicate lesions that are visually restored closer to the ground truth by our method. The last row depicts the difference map between the restored images and the ground truth.}
	\label{fig:prostate_figure_qualitative}
\end{figure}

We compared our proposed Res-SRDiff model against Bicubic, CycleGAN, Pix2pix, SPSR, I\textsuperscript{2}SR, and MT-DDPM. Our proposed method was able to restore axial T2w pelvic images with improved fidelity to the HR ground truth, as shown in \figurename~\ref{fig:prostate_figure_qualitative}. Although the Pix2pix and SPSR methods  successfully restored HR images that were globally similar to the ground truth, our method better restored the lesion, as indicated by the red arrow in the second row of \figurename~\ref{fig:prostate_figure_qualitative}. These findings are further confirmed by the difference maps shown in the third row of \figurename~\ref{fig:prostate_figure_qualitative}, where our method exhibits the smallest residual error compared with the other methods.

We conducted a comparative evaluation of our proposed Res-SRDiff model against Bicubic, CycleGAN, Pix2pix, SPSR, I\textsuperscript{2}SR, and MT-DDPM. Our method restored axial T2w pelvic images with enhanced fidelity to the HR ground truth, as illustrated in \figurename~\ref{fig:prostate_figure_qualitative}. Although the Pix2pix and SPSR methods generated HR images that were globally similar to the ground truth, our approach more effectively restored the lesion, as indicated by the red arrow in the second row of \figurename~\ref{fig:prostate_figure_qualitative}. This observation is further confirmed by the difference maps shown in the third row, where our method exhibits the smallest residual error relative to the alternative approaches.

\begin{figure}[tbhp]
	\centering
	\includegraphics[width=\textwidth]{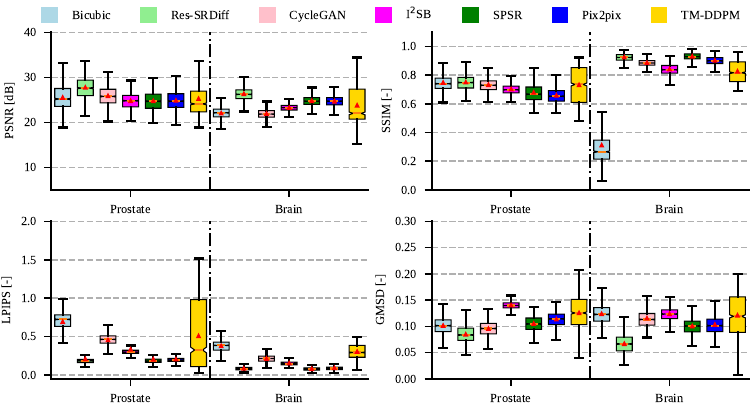}
	\caption{The boxplot of the quantitative metrics is illustrated for the T2w pelvic and T1 map brain MRI. }
	\label{fig:boxplot_rev1}
\end{figure}

In a further analysis, the evaluation time of our proposed method ($0.95 \pm 0.74$ seconds per slice) remained considerably lower than of I\textsuperscript{2}SR ($38.32 \pm 25.74$ seconds per slice) and that of MT-DDPM ($20.66 \pm 14.00$ seconds per slice). The quantitative metrics failed the Shapiro-Wilk normality test (with \textit{p}-values $\ll 0.001$); thus, we performed non-parametric Kruskal-Wallis and Dunn's tests. The Kruskal-Wallis test yielded \textit{p}-values $\ll 0.0001$, indicating that the differences between the methods were statistically significant for all quantitative metrics. Specifically, our method achieved the highest PSNR ($27.72 \pm 2.26$) and the lowest GMSD ($0.08 \pm 0.02$). Although our method, on average, achieved the second-best LPIPS after Pix2pix and SPRS, the differences were not statistically significant with \textit{p} $= 0.17$ and \textit{p} $= 0.31$, respectively. \tablename~\ref{tab:qmetrics} summarizes the quantitative metrics and indicates whether the differences are statistically significant. Furthermore, \figurename~\ref{fig:boxplot_rev1} illustrates the boxplot of the quantitative metrics where vertical orange lines and red rectangular markers show the median and average values.

\begin{table}[t]
	\caption{Quantitative comparison of super-resolution models on two datasets: Axial T2w pelvic MRI and 7T brain T1 MP2RAGE maps. Results are presented as mean $\pm$ standard deviation for our proposed Res-SRDiff and comparative models. Bold values highlight the best-performing results, while underlined values indicate the second-best performance. Arrows indicate the direction of better results.}
	\label{tab:qmetrics}
	\resizebox{\columnwidth}{!}{%
		\begin{tabular}{llllllllll}
			\hline
			& \multicolumn{4}{c}{Pelvic T2w MRI}                           &  & \multicolumn{4}{c}{7T brain T1 MP2RAGE map}                    \\ \cline{2-5} \cline{7-10} 
			Models     & PSNR [dB] $ \uparrow $     & SSIM [-]  $ \uparrow $    & GMSD [-]  $ \downarrow $    & LPIPS [-]  $ \downarrow $    &  & PSNR [dB]   $ \uparrow $   & SSIM [-] $ \uparrow $     & GMSD [-]  $ \downarrow $   & LPIPS [-]    $ \downarrow $  \\ \hline
			Bicubic    & $25.47 _{\pm 2.61}$ & $\mathbf{0.75 _{\pm 0.06}} ^*$ & $\underline{0.10 _{\pm 0.02}}$ & $0.69 _{\pm 0.15}$ &  & $22.00 _{\pm 1.37}$ & $0.31 _{\pm 0.16}$ & $0.12 _{\pm 0.02}$ & $0.38 _{\pm 0.07}$ \\
			cycleGAN   & $\underline{25.84 _{\pm 1.96}}$ & $\underline{0.73 _{\pm 0.05}}$ & $\underline{0.10 _{\pm 0.01}}$ & $0.45 _{\pm 0.10}$ &  & $21.89 _{\pm 1.09}$ & $0.86 _{\pm 0.02}$ & $0.12 _{\pm 0.02}$ & $0.21 _{\pm 0.05}$ \\
			Pix2pix    & $24.83 _{\pm 2.09}$ & $0.66 _{\pm 0.05}$ & $0.11 _{\pm 0.01}$ & $\mathbf{0.20 _{\pm 0.05}}^*$ &  & ${24.63 _{\pm 1.32}}$ & ${0.90 _{\pm 0.03}}$ & $\underline{0.10 _{\pm 0.02}}$ & $\underline{0.09 _{\pm 0.04}}^*$ \\
			TM-DDPM    & $25.12 _{\pm 4.46}$ & $\underline{0.73 _{\pm 0.16}}$ & $0.13 _{\pm 0.04}$ & $0.51 _{\pm 0.49}$ &  & $23.22 _{\pm 5.02}$ & $0.85 _{\pm 0.13}$ & $0.12 _{\pm 0.05}$ & $0.25 _{\pm 0.10}$ \\
			SPSR    & $24.74_{\pm 1.96}$ & $0.68_{\pm 0.07}$ & $0.11_{\pm 0.01}$ & $\mathbf{0.20_{\pm 0.09}}^*$ &  & $\underline{24.76_{\pm 1.12}}$ & $\mathbf{0.93_{\pm 0.02}}$ & $\underline{0.10_{\pm 0.01}}$ & $\mathbf{0.08_{\pm 0.02}}$ \\
			I\textsuperscript{2}SB    & $24.74_{\pm 1.60}$ & $0.70_{\pm 0.04}$ & $0.14_{\pm 0.01}$ & $0.33_{\pm 0.13}$ &  & $23.22_{\pm 0.98}$ & $0.84_{\pm 0.04}$ & $0.12_{\pm 0.01}$ & $0.15_{\pm 0.03}$ \\
			Res-SRDiff & $\mathbf{27.72 _{\pm 2.26}}$ & $\mathbf{0.75 _{\pm 0.05}}$ & $\mathbf{0.08 _{\pm 0.02}}$ & $\underline{0.21 _{\pm 0.11}}$ &  & $\mathbf{26.28 _{\pm 1.41}}$ & $\underline{0.92 _{\pm 0.03}}$ & $\mathbf{0.07 _{\pm 0.02}}$ & $\mathbf{0.08 _{\pm 0.02}}$ \\ \hline
			\multicolumn{10}{l}{$ ^* $ {\footnotesize denotes results that are not statistically significant based on the multi-comparison test (\textit{p}-value $> 0.05$).}}
		\end{tabular}%
	}
\end{table}

The Shapiro-Wilk test revealed that the Likert scores were not normally distributed (\textit{p} $\ll 0.001$) across all evaluated methods. Our proposed method achieved an average score of $4.80 \pm 0.40$ (95\% CI 4.65, 4.88), surpassing the second-best method, TM-DDPM, which obtained an average score of $4.33 \pm 0.68$ (95\% CI 4.12, 4.49). Nonetheless, the difference was not statistically significant (\textit{p} $\ll 0.16$). Furthermore, both of these methods significantly outperformed the third-best method, Pix2pix, which scored $3.67 \pm 0.55$ (95\% CI 3.49, 3.80) (\textit{p} $\ll 0.001$).

\subsection{Ablation study}
The ablation study was conducted to evaluate the contribution of the Swin Transformer block, which replaces the attention layer, and U-net without Swin Transformer block to the overall performance of the Res-SRDiff method. The quantitative results are presented in \tablename~\ref{tab:ablation}. The Shapiro test indicated that PSNR and SSIM for brain dataset were not normally distribution with \textit{p}-values of $0.046$ and $\ll 10^{-4}$, and all the metrics for pelvis dataset were not normally distributed with \textit{p}-values $\ll 10^{-4}$. The paired \textit{t}-test or Wilcoxon signed rank test results, according to the data distribution, are summarized in \tablename~\ref{tab:ablation}.

\begin{table}[tbhp]
	\caption{The Ablation study results are summarized. The best results are written in bold. The absolute percentage changes are written inside a paranthesis where \textcolor{red}{red} color and \textcolor{blue}{blue} color indicate percentage of improvement and reduction of performance, respectively, of the Res-SRDiff model compare with its variation without using Swin Transformer block.}
	\label{tab:ablation}
	\resizebox{\textwidth}{!}{%
		\begin{tabular}{lccllll}\hline
			& \multicolumn{2}{c}{Training scenario} &                       &                      &                      &                      \\\cdashline{2-3}
			Imaging region                                                                      & \multicolumn{1}{c}{w/o$^1$} & \multicolumn{1}{c}{w$^2$}           & PSNR [dB] $\uparrow$  & SSIM [-]    $\uparrow$         & GMSD [-]    $\downarrow$         & LIPIPS [-]       $\downarrow$       \\ \hline
			\multirow{2}{*}{Pelvis T2w MRI}                                                     &         $\bullet$           &      $\circ$            &  $26.92_{\pm 2.18} $       & $0.71_{\pm 0.05} $       & $0.10_{\pm 0.02}$        & $0.22_{\pm 0.10}$        \\
			&             $\circ$        &       $\bullet$           & $\mathbf{27.72} _{\mathbf{\pm 2.26}\textcolor{red}{(2.97)}}$ & $\textbf{0.75}_{\pm \textbf{0.05}\textcolor{red}{(5.63)}}$ & $\textbf{0.08}_{\pm \textbf{0.02}\textcolor{red}{(20.00)}}$ &$ \textbf{0.21}_{\pm \textbf{0.11}\textcolor{red}{(4.55)}}$  \\\cdashline{1-7}
			\multirow{2}{*}{\begin{tabular}[c]{@{}l@{}}7T brain T1 \\ MP2RAGE map\end{tabular}} &       $\bullet$             &        $\circ$           & $ \textbf{26.41} _{\pm \textbf{1.42}}$$ ^* $        & $0.90_{\pm 0.02}$        & $0.10_{\pm 0.02}$        & $0.10_{\pm 0.02} $       \\
			&          $\circ$           &           $\bullet$       & ${26.28}_{\pm {1.41}\textcolor{blue}{(0.49)}}$  & $\textbf{0.92}_{\pm \textbf{0.03}\textcolor{red}{(2.22)}} $ & $\textbf{0.09}_{\pm \textbf{0.02}\textcolor{red}{(10.00)}}$ & $\textbf{0.08}_{\pm \textbf{0.02}\textcolor{red}{(20.00)}}$ \\ \hline
			\multicolumn{7}{l}{\begin{tabular}[c]{@{}l@{}}$^1$Trained without using Swin Transformer block.\\ $^2$Trained using Swin Transformer block (Res-SRDiff).\\ 
					*Denotes results that are not statistically significant(\textit{p}-value $> 0.05$).
			\end{tabular}} 
		\end{tabular}%
	}
\end{table}          

Res-SRDiff with Swin Transformer block achieved the superior performance in reconstructing high-resolution pelvis T2w MRI compared with the U-net for all quantitative metrics. Specifically, Swin Transformer block increases PSNR and SSIM about $ 2.97\% $ and $ 5.63\% $, and reduces GMSD and LPIPS $ 20\% $ and $ 4.55\% $, where all the differences were statistically significant different with \textit{p}-values $ \ll 10^{-4} $. 

Similarly, Res-SRDiff with Swin Transformer block achieved the superior performance in reconstructing high-resolution brain T1 map compared with the U-net for SSIM, GMSD, and LPIPS metrics. Specifically, using Swin Transformer block increased SSIM $ 2.22 \% $ and reduced GMSD and LPIPS $ 10\% $ and $ 20\% $. However, Res-SRDiff that did not use Swin Transformer block achieved a lower PSNR  value. All the differences were statistically significant with \textit{p}-values $ \ll 10^{-4} $, except PSNR metric with \textit{p} $ =0.09 $. 

Res-SRDiff using the Swin Transformer block could successfully preserved the fine-details better than when it uses a U-net without Swin Transformer block indicated by red and white arrows in \figurename~6. Although the global difference between the two scenrios are small in reconstructing the pelvis T2w images, the Res-SRDiff with Swin Transformer block was able to reconstruct images with sharper details indicated by yellow arrow in \figurename~6.

\begin{figure}[h]
	\centering
	\includegraphics[width=\textwidth]{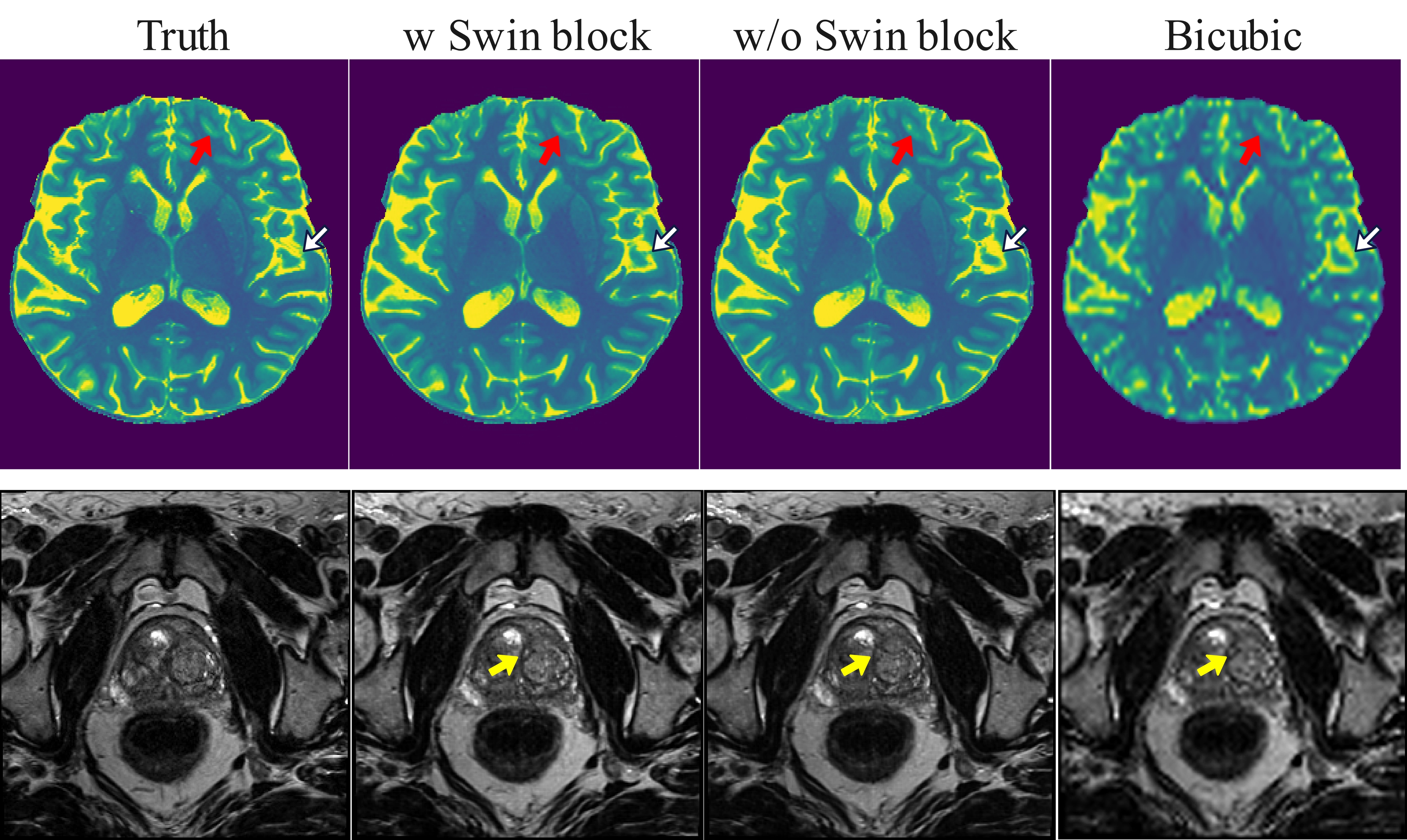}
	\caption{The ablation results to reconstruct the brain and pelvis MRI images are illustrated for scenarios where the Res-SRDiff was trained with Swin Transformer block (w Swin block) and without Swin Transformer block (w/o Swin block).}
	\label{fig:ablation_brainProstate}
\end{figure}

\begin{figure}[h]
	\centering
	\includegraphics[width=.8\textwidth]{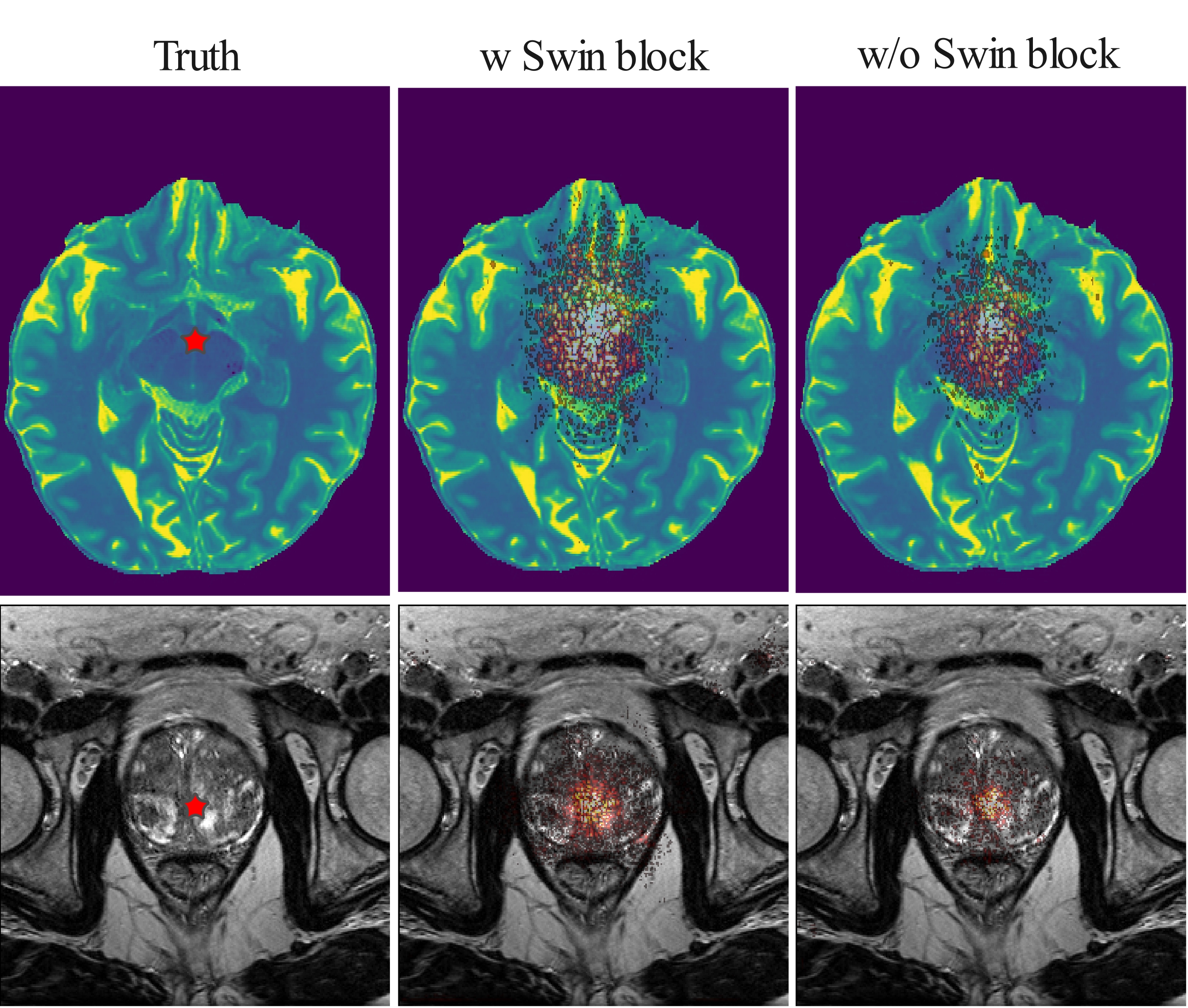}
	\caption{The influence of pixel neighborhood to reconstruct the region shown by red stars is shown for the scenarios where the Res-SRDiff was trained with (w Swin block) and without (w/o Swin block) Swin Transformer blocks. Wider distributions indicate the involvement of more pixels that might help to reconstruct the region with higher fidelity.}
	\label{fig:grad_map}
\end{figure}

In addition, we compared the local attribution map (LAM) to visualize the influence of the neighborhood pixels to reconstruct a shown with star region in \figurename~\ref{fig:grad_map}. By leveraging broader range of information, the Res-SRDiff with Swin Transformer block achieved improved results. Compared with the Res-SRDiff without Swin Transformer block, the proposed method with Swin Transformer block leverage wider information to reconstruct the given region for both brain and pelvis MRI (see \figurename~\ref{fig:grad_map}).

\section{Discussion}\label{sec:discussion}

MRI remains one of the most versatile modalities in both clinical practice and research due to its excellent soft-tissue contrast and ability to generate multiple image contrasts without ionizing radiation. However, the inherently long acquisition times can lead to patient discomfort and motion artifacts~\cite{Mojtaba_2024_marcdpm}, often forcing a trade‐off between spatial resolution and acquisition efficiency. One of the easiest approaches to mitigate these challenges is to increase the voxel size, but this can adversely affect the diagnostic quality~\cite{QIMS111763} by introducing partial volume effects.

In this study, we introduced \textbf{Res-SRDiff}, an efficient probabilistic diffusion model designed to reconstruct HR MRI images from LR inputs. By leveraging the residual error between the LR and HR images in the forward diffusion process, our approach shifts the HR image distribution toward that of the LR images. This enables the reverse process using a NN to accurately recover fine image details in only four sampling steps, markedly reducing the reconstruction time to under \textbf{one second per slice} compared with conventional diffusion models, which may require up to 20 seconds per slice.

Our experiments on both brain T1 maps and pelvic T2w images demonstrate that Res-SRDiff not only improves computational efficiency but also preserves critical anatomical details. For the brain T1 maps, qualitative assessments (as indicated by the white and red arrows in \figurename~\ref{fig:brain_figure_qualitative}) reveal that our method recovers fine structures with smaller residual errors compared to competing models. Quantitatively, our approach consistently achieved the highest PSNR and lowest GMSD, with statistically significant improvements (\textit{p} $ \ll 0.05 $). Moreover, the small standard deviation observed across test samples suggests that incorporating the residual error $ e_0 $ contributes to a more stable and robust reconstruction process.

Similarly, in the pelvic T2w images, Res-SRDiff successfully reconstructs HR images with improved lesion depiction. Unlike the TM-DDPM method--which tended to exaggerate lesion sizes, possibly due to its progressive sampling process--our method maintained more anatomically accurate representations while also exhibiting lower residual errors. These findings align with the previous study that reported that DDPMs tend to generate blurry images~\cite{https://doi.org/10.1002/mp.17258}. The consistency of these results across both datasets underscores the advantage of integrating residual error information into the diffusion process.

The application of our SR approach offers substantial clinical benefits by enhancing lesion diagnosis and contouring in both brain and prostate imaging. For the brain T1 MP2RAGE maps, our method preserves fine anatomical details and reduces residual errors, which is critical for differentiating subtle lesions such as those associated with tumors, demyelinating diseases, or stroke. Similarly, the improved depiction of lesion morphology in pelvic T2w images, particularly in areas where traditional methods tend to exaggerate lesion boundaries, underscores the potential for more accurate identification and delineation of prostate lesions. In both cases, the enhanced image quality may not only support more precise treatment planning by reducing diagnostic and contouring uncertainty but also contribute to decreased acquisition times, thereby minimizing patient discomfort and motion artifacts. This comprehensive improvement in image quality across multiple anatomical regions highlights the clinical relevance and robustness of the Res-SRDiff method, paving the way for its integration into routine diagnostic and treatment workflows.

Looking forward, several promising research avenues arise from our work. Expanding the Res-SRDiff framework to include other imaging modalities and incorporating it into real-time clinical workflows could remarkably enhance its effectiveness. Furthermore, refining the diffusion process by incorporating adaptive noise scheduling~\cite{lee2024ant} or aligning measurement gradient with local manifold structure of the Res-SRDiff diffusion state~\cite{zirvi2025diffusion} has the potential to further improve image quality.

This study presents limitations that merit attention. Although the datasets comprise 3D volumes with resolution reduced across all three dimensions, the Res-SRDiff framework is implemented as a 2D super-resolution method that processes individual slices independently. This approach capitalizes on the efficiency of 2D convolutional architectures and simplifies both training and inference. However, because each slice is enhanced separately, there is no explicit mechanism in the current model to guarantee inter-slice continuity. As such, although improved spatial resolution and lesion delineation are achieved on a per-slice basis, the volumetric consistency across slices remains an open challenge. Future work will aim to address this limitation, potentially by incorporating 3D super-resolution architectures or additional continuity constraints to ensure robust inter-slice coherence in fully volumetric reconstructions. In addition, we acknowledge that the SR process may inadvertently alter or omit very fine anatomical structures (pink arrows shown in \figurename~\ref{fig:brain_figure_qualitative}). While such alterations are rare, they could have clinical implications, and future improvements should focus on incorporating structure-preserving mechanisms or uncertainty estimation to better safeguard clinically relevant features.

\section{Conclusions}\label{sec:conclusion}

The proposed Res-SRDiff marks a substantial advancement in the creation of efficient diffusion-based super-resolution models for MRI. By minimizing the number of necessary sampling steps and utilizing residual error information, our approach achieves superior image restoration performance while ensuring both computational efficiency and consistency across a range of datasets.

Res-SRDiff provides a highly efficient and precise framework for MRI super-resolution, offering a notable reduction in computational time while maintaining or even exceeding the image quality of state-of-the-art methods. The integration of residual error shifting within the diffusion process signifies a meaningful step forward in medical image reconstruction, with potential implications for accelerating high-quality imaging in both clinical workflows and research applications.

\section*{Conflicts of interest}
There are no conflicts of interest declared by the authors.

\section*{Acknowledgment}
This research is supported in part by the National Institutes of Health under Award Numbers R56EB033332, R01DE033512, and R01CA272991.

\section*{Data availability}
The ProstateX data is publicly available at the TCIA portal (\url{https://www.cancerimagingarchive.net/analysis-result/prostatex-seg-hires/}). Our institutional data cannot be made publicly available upon publication because they contain sensitive personal information.



\begin{thebibliography}{100}
	
	\bibitem{MARQUES20101271}
	José~P. Marques, Tobias Kober, Gunnar Krueger, Wietske {van der Zwaag},
	Pierre-François {Van de Moortele}, and Rolf Gruetter.
	\newblock Mp2rage, a self bias-field corrected sequence for improved
	segmentation and t1-mapping at high field.
	\newblock {\em NeuroImage}, 49(2):1271--1281, 2010.
	
	\bibitem{doi:10.1148/radiol.2016141802}
	Erik~B. Schelbert and Daniel~R. Messroghli.
	\newblock State of the art: Clinical applications of cardiac t1 mapping.
	\newblock {\em Radiology}, 278(3):658--676, 2016.
	\newblock PMID: 26885733.
	
	\bibitem{doi:10.1016/j.jcmg.2015.11.005}
	Andrew~J. Taylor, Michael Salerno, Rohan Dharmakumar, and Michael
	Jerosch-Herold.
	\newblock T1 mapping.
	\newblock {\em JACC: Cardiovascular Imaging}, 9(1):67--81, 2016.
	
	\bibitem{doi:10.1148/radiol.12120777}
	Eric~R. Muir, Yi~Zhang, Oscar San Emeterio~Nateras, Qi~Peng, and Timothy~Q.
	Duong.
	\newblock Human vitreous: Mr imaging of oxygen partial pressure.
	\newblock {\em Radiology}, 266(3):905--911, 2013.
	\newblock PMID: 23220896.
	
	\bibitem{EPEL2019977}
	Boris Epel, Matthew~C. Maggio, Eugene~D. Barth, Richard~C. Miller, Charles~A.
	Pelizzari, Martyna Krzykawska-Serda, Subramanian~V. Sundramoorthy, Bulent
	Aydogan, Ralph~R. Weichselbaum, Victor~M. Tormyshev, and Howard~J. Halpern.
	\newblock Oxygen-guided radiation therapy.
	\newblock {\em International Journal of Radiation Oncology*Biology*Physics},
	103(4):977--984, 2019.
	
	\bibitem{https://doi.org/10.1002/mp.17353}
	Claire Keun~Sun Park, Noah~Stanley Warner, Evangelia Kaza, and Atchar
	Sudhyadhom.
	\newblock Optimization and validation of low-field mp2rage t1 mapping on 0.35t
	mr-linac: Toward adaptive dose painting with hypoxia biomarkers.
	\newblock {\em Medical Physics}, 51(11):8124--8140, 2024.
	
	\bibitem{https://doi.org/10.1002/mp.14196}
	Yabo Fu, Yang Lei, Tonghe Wang, Sibo Tian, Pretesh Patel, Ashesh~B. Jani,
	Walter~J. Curran, Tian Liu, and Xiaofeng Yang.
	\newblock Pelvic multi-organ segmentation on cone-beam ct for prostate adaptive
	radiotherapy.
	\newblock {\em Medical Physics}, 47(8):3415--3422, 2020.
	
	\bibitem{Safari_2024_MAUDGAN}
	Mojtaba Safari, Xiaofeng Yang, Chih-Wei Chang, Richard L~J Qiu, Ali Fatemi, and
	Louis Archambault.
	\newblock Unsupervised mri motion artifact disentanglement: introducing
	maudgan.
	\newblock {\em Physics in Medicine \& Biology}, 69(11):115057, may 2024.
	
	\bibitem{LEPCHA2023230}
	Dawa~Chyophel Lepcha, Bhawna Goyal, Ayush Dogra, and Vishal Goyal.
	\newblock Image super-resolution: A comprehensive review, recent trends,
	challenges and applications.
	\newblock {\em Information Fusion}, 91:230--260, 2023.
	
	\bibitem{10.1007/978-3-540-79490-5_8}
	Xin Zhang, Edmund~Y. Lam, Ed~X. Wu, and Kenneth K.~Y. Wong.
	\newblock Application of tikhonov regularization to super-resolution
	reconstruction of brain mri images.
	\newblock In Xiaohong Gao, Henning M{\"u}ller, Martin~J. Loomes, Richard
	Comley, and Shuqian Luo, editors, {\em Medical Imaging and Informatics},
	pages 51--56, Berlin, Heidelberg, 2008. Springer Berlin Heidelberg.
	
	\bibitem{6392274}
	Weisheng Dong, Lei Zhang, Guangming Shi, and Xin Li.
	\newblock Nonlocally centralized sparse representation for image restoration.
	\newblock {\em IEEE Transactions on Image Processing}, 22(4):1620--1630, 2013.
	
	\bibitem{4770145}
	CÉdric Vonesch and Michael Unser.
	\newblock A fast multilevel algorithm for wavelet-regularized image
	restoration.
	\newblock {\em IEEE Transactions on Image Processing}, 18(3):509--523, 2009.
	
	\bibitem{5193008}
	Shantanu~H. Joshi, Antonio Marquina, Stanley~J. Osher, Ivo Dinov, John~D.
	Van~Horn, and Arthur~W. Toga.
	\newblock Mri resolution enhancement using total variation regularization.
	\newblock In {\em 2009 IEEE International Symposium on Biomedical Imaging: From
		Nano to Macro}, pages 161--164, 2009.
	
	\bibitem{https://doi.org/10.1002/mp.17675}
	Mojtaba Safari, Zach Eidex, Shaoyan Pan, Richard L.~J. Qiu, and Xiaofeng Yang.
	\newblock Self-supervised adversarial diffusion models for fast mri
	reconstruction.
	\newblock {\em Medical Physics}, n/a(n/a).
	
	\bibitem{9703109}
	Yutong Chen, Carola-Bibiane Schönlieb, Pietro Liò, Tim Leiner, Pier~Luigi
	Dragotti, Ge~Wang, Daniel Rueckert, David Firmin, and Guang Yang.
	\newblock Ai-based reconstruction for fast mri—a systematic review and
	meta-analysis.
	\newblock {\em Proceedings of the IEEE}, 110(2):224--245, 2022.
	
	\bibitem{9340274}
	Pranaba~K. Mishro, Sanjay Agrawal, Rutuparna Panda, and Ajith Abraham.
	\newblock A survey on state-of-the-art denoising techniques for brain magnetic
	resonance images.
	\newblock {\em IEEE Reviews in Biomedical Engineering}, 15:184--199, 2022.
	
	\bibitem{YI2019101552}
	Xin Yi, Ekta Walia, and Paul Babyn.
	\newblock Generative adversarial network in medical imaging: A review.
	\newblock {\em Medical Image Analysis}, 58:101552, 2019.
	
	\bibitem{10738507}
	Longguang Wang, Yulan Guo, Yingqian Wang, Xiaoyu Dong, Qingyu Xu, Jungang Yang,
	and Wei An.
	\newblock Unsupervised degradation representation learning for unpaired
	restoration of images and point clouds.
	\newblock {\em IEEE Transactions on Pattern Analysis and Machine Intelligence},
	47(1):1--18, 2025.
	
	\bibitem{Ma_2020_CVPR}
	Cheng Ma, Yongming Rao, Yean Cheng, Ce~Chen, Jiwen Lu, and Jie Zhou.
	\newblock Structure-preserving super resolution with gradient guidance.
	\newblock In {\em Proceedings of the IEEE/CVF Conference on Computer Vision and
		Pattern Recognition (CVPR)}, June 2020.
	
	\bibitem{10.1117/12.3002863}
	Mojtaba Safari, Xiaofeng Yang, and Ali Fatemi.
	\newblock {MRI data consistency guided conditional diffusion probabilistic
		model for MR imaging acceleration}.
	\newblock In Barjor~S. Gimi and Andrzej Krol, editors, {\em Medical Imaging
		2024: Clinical and Biomedical Imaging}, volume 12930, page 129300R.
	International Society for Optics and Photonics, SPIE, 2024.
	
	\bibitem{Pan_2023}
	Shaoyan Pan, Tonghe Wang, Richard L~J Qiu, Marian Axente, Chih-Wei Chang, Junbo
	Peng, Ashish~B Patel, Joseph Shelton, Sagar~A Patel, Justin Roper, and
	Xiaofeng Yang.
	\newblock 2d medical image synthesis using transformer-based denoising
	diffusion probabilistic model.
	\newblock {\em Physics in Medicine \& Biology}, 68(10):105004, may 2023.
	
	\bibitem{10.1117/12.3047506}
	Shaoyan Pan, Zach Eidex, Mojtaba Safari, Richard Qiu, and Xiaofeng Yang.
	\newblock {Cycle-guided denoising diffusion probability model for 3D
		cross-modality MRI synthesis}.
	\newblock In Barjor~S. Gimi and Andrzej Krol, editors, {\em Medical Imaging
		2025: Clinical and Biomedical Imaging}, volume 13410, page 134101W.
	International Society for Optics and Photonics, SPIE, 2025.
	
	\bibitem{Chang_2024}
	Chih-Wei Chang, Junbo Peng, Mojtaba Safari, Elahheh Salari, Shaoyan Pan, Justin
	Roper, Richard L~J Qiu, Yuan Gao, Hui-Kuo Shu, Hui Mao, and Xiaofeng Yang.
	\newblock High-resolution mri synthesis using a data-driven framework with
	denoising diffusion probabilistic modeling.
	\newblock {\em Physics in Medicine \& Biology}, 69(4):045001, feb 2024.
	
	\bibitem{10737883}
	Brian~B. Moser, Arundhati~S. Shanbhag, Federico Raue, Stanislav Frolov,
	Sebastian Palacio, and Andreas Dengel.
	\newblock Diffusion models, image super-resolution, and everything: A survey.
	\newblock {\em IEEE Transactions on Neural Networks and Learning Systems},
	pages 1--21, 2024.
	
	\bibitem{yue2023resshift}
	Jianyi~Wang Zongsheng~Yue and Chen~Change Loy.
	\newblock Resshift: Efficient diffusion model for image super-resolution by
	residual shifting.
	\newblock In {\em Advances in Neural Information Processing Systems (NeurIPS)},
	2023.
	
	\bibitem{10681246}
	Zongsheng Yue, Jianyi Wang, and Chen~Change Loy.
	\newblock Efficient diffusion model for image restoration by residual shifting.
	\newblock {\em IEEE Transactions on Pattern Analysis and Machine Intelligence},
	47(1):116--130, 2025.
	
	\bibitem{liu2023schrodinger}
	Guan-Horng Liu, Arash Vahdat, De-An Huang, Evangelos Theodorou, Weili Nie, and
	Anima Anandkumar.
	\newblock I\^2sb: Image-to-image schr\"{o}dinger bridge.
	\newblock In {\em International Conference on Machine Learning (ICML)}, July
	2023.
	
	\bibitem{10720924}
	Kai Zhao, Kaifeng Pang, Alex Ling~Yu Hung, Haoxin Zheng, Ran Yan, and Kyunghyun
	Sung.
	\newblock Mri super-resolution with partial diffusion models.
	\newblock {\em IEEE Transactions on Medical Imaging}, pages 1--1, 2024.
	
	\bibitem{Wang_2024_CVPR}
	Yufei Wang, Wenhan Yang, Xinyuan Chen, Yaohui Wang, Lanqing Guo, Lap-Pui Chau,
	Ziwei Liu, Yu~Qiao, Alex~C. Kot, and Bihan Wen.
	\newblock Sinsr: Diffusion-based image super-resolution in a single step.
	\newblock In {\em Proceedings of the IEEE/CVF Conference on Computer Vision and
		Pattern Recognition (CVPR)}, pages 25796--25805, June 2024.
	
	\bibitem{pmlr-v37-sohl-dickstein15}
	Jascha Sohl-Dickstein, Eric Weiss, Niru Maheswaranathan, and Surya Ganguli.
	\newblock Deep unsupervised learning using nonequilibrium thermodynamics.
	\newblock In Francis Bach and David Blei, editors, {\em Proceedings of the 32nd
		International Conference on Machine Learning}, volume~37 of {\em Proceedings
		of Machine Learning Research}, pages 2256--2265, Lille, France, 07--09 Jul
	2015. PMLR.
	
	\bibitem{song2021scorebased}
	Yang Song, Jascha Sohl-Dickstein, Diederik~P Kingma, Abhishek Kumar, Stefano
	Ermon, and Ben Poole.
	\newblock Score-based generative modeling through stochastic differential
	equations.
	\newblock In {\em International Conference on Learning Representations}, 2021.
	
	\bibitem{10.5555/3495724.3496298}
	Jonathan Ho, Ajay Jain, and Pieter Abbeel.
	\newblock Denoising diffusion probabilistic models.
	\newblock In {\em Proceedings of the 34th International Conference on Neural
		Information Processing Systems}, NIPS '20, Red Hook, NY, USA, 2020. Curran
	Associates Inc.
	
	\bibitem{luo2022understandingdiffusionmodelsunified}
	Calvin Luo.
	\newblock Understanding diffusion models: A unified perspective, 2022.
	
	\bibitem{pml1Book}
	Kevin~P. Murphy.
	\newblock {\em Probabilistic Machine Learning: An Introduction}.
	\newblock MIT Press, 2022.
	
	\bibitem{Zhang_2018_CVPR}
	Richard Zhang, Phillip Isola, Alexei~A. Efros, Eli Shechtman, and Oliver Wang.
	\newblock The unreasonable effectiveness of deep features as a perceptual
	metric.
	\newblock In {\em Proceedings of the IEEE Conference on Computer Vision and
		Pattern Recognition (CVPR)}, June 2018.
	
	\bibitem{liu2021varianceadaptivelearningrate}
	Liyuan Liu, Haoming Jiang, Pengcheng He, Weizhu Chen, Xiaodong Liu, Jianfeng
	Gao, and Jiawei Han.
	\newblock On the variance of the adaptive learning rate and beyond, 2021.
	
	\bibitem{loshchilov2017sgdrstochasticgradientdescent}
	Ilya Loshchilov and Frank Hutter.
	\newblock Sgdr: Stochastic gradient descent with warm restarts, 2017.
	
	\bibitem{Middlebrooks_2024_data}
	Erik~H. Middlebrooks, Vishal Patel, Xiangzhi Zhou, Sina Straub, John~V.
	Murray~Jr., Amit~K. Agarwal, Lela Okromelidze, Rahul~B. Singh, Alfonso~S.
	Lopez~Chiriboga, Erin~M. Westerhold, Vivek Gupta, Sukhwinder Johnny~Singh
	Sandhu, Iris~V. Marin~Collazo, and Shengzhen Tao.
	\newblock 7 t lesion-attenuated magnetization-prepared gradient echo
	acquisition for detection of posterior fossa demyelinating lesions in
	multiple sclerosis.
	\newblock {\em Investigative Radiology}, 59(7), 2024.
	
	\bibitem{10.1117/1.JMI.5.4.044501}
	Samuel~G. Armato, Henkjan Huisman, Karen Drukker, Lubomir Hadjiiski, Justin~S.
	Kirby, Nicholas Petrick, George Redmond, Maryellen~L. Giger, Kenny Cha, Artem
	Mamonov, Jayashree Kalpathy-Cramer, and Keyvan Farahani.
	\newblock {PROSTATEx Challenges for computerized classification of prostate
		lesions from multiparametric magnetic resonance images}.
	\newblock {\em Journal of Medical Imaging}, 5(4):044501, 2018.
	
	\bibitem{https://doi.org/10.1002/hbm.10062}
	Stephen~M. Smith.
	\newblock Fast robust automated brain extraction.
	\newblock {\em Human Brain Mapping}, 17(3):143--155, 2002.
	
	\bibitem{Yaniv2018}
	Ziv Yaniv, Bradley~C. Lowekamp, Hans~J. Johnson, and Richard Beare.
	\newblock Simpleitk image-analysis notebooks: a collaborative environment for
	education and reproducible research.
	\newblock {\em Journal of Digital Imaging}, 31(3):290--303, Jun 2018.
	
	\bibitem{isola2017image}
	Phillip Isola, Jun-Yan Zhu, Tinghui Zhou, and Alexei~A Efros.
	\newblock Image-to-image translation with conditional adversarial networks.
	\newblock In {\em Computer Vision and Pattern Recognition (CVPR), 2017 IEEE
		Conference on}, 2017.
	
	\bibitem{CycleGAN2017}
	Jun-Yan Zhu, Taesung Park, Phillip Isola, and Alexei~A Efros.
	\newblock Unpaired image-to-image translation using cycle-consistent
	adversarial networks.
	\newblock In {\em Computer Vision (ICCV), 2017 IEEE International Conference
		on}, 2017.
	
	\bibitem{1284395}
	Zhou Wang, A.C. Bovik, H.R. Sheikh, and E.P. Simoncelli.
	\newblock Image quality assessment: from error visibility to structural
	similarity.
	\newblock {\em IEEE Transactions on Image Processing}, 13(4):600--612, 2004.
	
	\bibitem{6678238}
	Wufeng Xue, Lei Zhang, Xuanqin Mou, and Alan~C. Bovik.
	\newblock Gradient magnitude similarity deviation: A highly efficient
	perceptual image quality index.
	\newblock {\em IEEE Transactions on Image Processing}, 23(2):684--695, 2014.
	
	\bibitem{1576816}
	H.R. Sheikh and A.C. Bovik.
	\newblock Image information and visual quality.
	\newblock {\em IEEE Transactions on Image Processing}, 15(2):430--444, 2006.
	
	\bibitem{Safari2023_medfusiongan}
	Mojtaba Safari, Ali Fatemi, and Louis Archambault.
	\newblock Medfusiongan: multimodal medical image fusion using an unsupervised
	deep generative adversarial network.
	\newblock {\em BMC Medical Imaging}, 23(1):203, Dec 2023.
	
	\bibitem{2020SciPy-NMeth}
	Pauli Virtanen, Ralf Gommers, Travis~E. Oliphant, Matt Haberland, Tyler Reddy,
	David Cournapeau, Evgeni Burovski, Pearu Peterson, Warren Weckesser, Jonathan
	Bright, St{\'e}fan~J. {van der Walt}, Matthew Brett, Joshua Wilson, K.~Jarrod
	Millman, Nikolay Mayorov, Andrew R.~J. Nelson, Eric Jones, Robert Kern, Eric
	Larson, C~J Carey, {\.I}lhan Polat, Yu~Feng, Eric~W. Moore, Jake
	{VanderPlas}, Denis Laxalde, Josef Perktold, Robert Cimrman, Ian Henriksen,
	E.~A. Quintero, Charles~R. Harris, Anne~M. Archibald, Ant{\^o}nio~H. Ribeiro,
	Fabian Pedregosa, Paul {van Mulbregt}, and {SciPy 1.0 Contributors}.
	\newblock {{SciPy} 1.0: Fundamental Algorithms for Scientific Computing in
		Python}.
	\newblock {\em Nature Methods}, 17:261--272, 2020.
	
	\bibitem{Mojtaba_2024_marcdpm}
	Mojtaba Safari, Xiaofeng Yang, Ali Fatemi, and Louis Archambault.
	\newblock Mri motion artifact reduction using a conditional diffusion
	probabilistic model (mar-cdpm).
	\newblock {\em Medical Physics}, 51(4):2598--2610, 2024.
	
	\bibitem{QIMS111763}
	Lijuan Mao, Xiaoling Zhang, Tingting Chen, Zhoulei Li, and Jianyong Yang.
	\newblock High-resolution reduced field-of-view diffusion-weighted magnetic
	resonance imaging in the diagnosis of cervical cancer.
	\newblock {\em Quantitative Imaging in Medicine and Surgery}, 13(6), 2023.
	
	\bibitem{https://doi.org/10.1002/mp.17258}
	Yuan Gao, Huiqiao Xie, Chih-Wei Chang, Junbo Peng, Shaoyan Pan, Richard L.~J.
	Qiu, Tonghe Wang, Beth Ghavidel, Justin Roper, Jun Zhou, and Xiaofeng Yang.
	\newblock Ct-based synthetic iodine map generation using conditional denoising
	diffusion probabilistic model.
	\newblock {\em Medical Physics}, 51(9):6246--6258, 2024.
	
	\bibitem{lee2024ant}
	Seunghan Lee, Kibok Lee, and Taeyoung Park.
	\newblock {ANT}: Adaptive noise schedule for time series diffusion models.
	\newblock In {\em The Thirty-eighth Annual Conference on Neural Information
		Processing Systems}, 2024.
	
	\bibitem{zirvi2025diffusion}
	Rayhan Zirvi, Bahareh Tolooshams, and Anima Anandkumar.
	\newblock Diffusion state-guided projected gradient for inverse problems.
	\newblock In {\em The Thirteenth International Conference on Learning
		Representations}, 2025.
	
\end{thebibliography}

\end{document}